\documentclass[11pt]{article}
\ifdefined\pdfminorversion\pdfminorversion=7\fi %

\newcommand{\method}{\textsc{OSVerify}\xspace}
\newcommand{\oss}{\texttt{oss}\xspace}
\newcommand{\mini}{\texttt{5mini}\xspace}

\usepackage{fix-cm}
\usepackage[utf8]{inputenc}
\usepackage[T1]{fontenc}
\usepackage{palatino}
\usepackage[letterpaper,textwidth=6.6in,textheight=8.6in,top=1in,
  headheight=12pt,headsep=25pt,footskip=30pt]{geometry}
\usepackage[numbers,sort&compress]{natbib}
\usepackage{etoolbox}
\usepackage{catchfile}

\usepackage{amsmath,amsfonts,bm}

\def\eqref#1{equation~\ref{#1}}

\def\1{\bm{1}}

\DeclareMathAlphabet{\mathsfit}{\encodingdefault}{\sfdefault}{m}{sl}
\SetMathAlphabet{\mathsfit}{bold}{\encodingdefault}{\sfdefault}{bx}{n}

\usepackage{amsmath}
\usepackage{amssymb}
\usepackage{booktabs}
\usepackage{enumitem}
\usepackage{etoc}
\usepackage{float}
\usepackage{graphicx}
\usepackage{subcaption}
\DeclareCaptionLabelFormat{panelwithfigure}{Figure~\thefigure#2}
\usepackage{wrapfig}
\usepackage{needspace}
\usepackage[colorlinks=true,breaklinks=true]{hyperref}
\usepackage{listings}
\usepackage[most]{tcolorbox}
\usepackage{microtype}
\usepackage{tikz}
\usepackage{cleveref}
\usepackage{xurl}
\usepackage{xspace}
\usepackage{fontawesome5}
\usepackage{xcolor}
\usepackage{colortbl}
\definecolor{osverifygreen}{HTML}{EDF2E9}
\definecolor{osverifylink}{HTML}{4F6B4F}
\definecolor{osverifyrose}{HTML}{F7ECEF}
\usetikzlibrary{arrows.meta,positioning}
\floatstyle{ruled}
\newfloat{algorithm}{tbp}{loa}
\floatname{algorithm}{Algorithm}

\makeatletter
\renewcommand{\normalsize}{%
  \@setfontsize\normalsize{11}{13}%
  \abovedisplayskip 6\p@ \@plus 2\p@ \@minus 5\p@
  \abovedisplayshortskip \z@ \@plus 3\p@
  \belowdisplayskip \abovedisplayskip
  \belowdisplayshortskip 4\p@ \@plus 3\p@ \@minus 3\p@}
\normalsize
\renewcommand{\small}{%
  \@setfontsize\small{9}{11}%
  \abovedisplayskip 10\p@ \@plus 1.5\p@ \@minus 4\p@
  \abovedisplayshortskip \z@ \@plus 2\p@
  \belowdisplayskip \abovedisplayskip
  \belowdisplayshortskip 3\p@ \@plus 2\p@ \@minus 2\p@}
\renewcommand{\footnotesize}{\@setfontsize\footnotesize{9}{10}}
\renewcommand{\scriptsize}{\@setfontsize\scriptsize{7}{8}}
\renewcommand{\tiny}{\@setfontsize\tiny{6}{7}}
\renewcommand{\large}{\@setfontsize\large{12}{14}}
\renewcommand{\Large}{\@setfontsize\Large{14}{16}}
\renewcommand{\LARGE}{\@setfontsize\LARGE{19}{24}}
\renewcommand{\huge}{\@setfontsize\huge{20}{23}}
\renewcommand{\Huge}{\@setfontsize\Huge{25}{28}}
\renewcommand{\section}{%
  \@startsection{section}{1}{\z@}%
  {-2.0ex \@plus -0.5ex \@minus -0.2ex}%
  {1.5ex \@plus 0.3ex \@minus 0.2ex}{\Large\bfseries\raggedright}}
\renewcommand{\subsection}{%
  \@startsection{subsection}{2}{\z@}%
  {-1.8ex \@plus -0.5ex \@minus -0.2ex}%
  {0.8ex \@plus 0.2ex}{\large\bfseries\raggedright}}
\renewcommand{\subsubsection}{%
  \@startsection{subsubsection}{3}{\z@}%
  {-1.5ex \@plus -0.5ex \@minus -0.2ex}%
  {0.5ex \@plus 0.2ex}{\normalsize\bfseries\raggedright}}
\renewcommand{\paragraph}{%
  \@startsection{paragraph}{4}{\z@}%
  {1.5ex \@plus 0.5ex \@minus 0.2ex}%
  {-1em}{\normalsize\bfseries\raggedright}}

\renewcommand{\maketitle}{\par
  \begingroup
    \renewcommand{\thefootnote}{\fnsymbol{footnote}}%
    \def\@makefnmark{\hbox to 4pt{\textcolor{osverifylink}{$^{\@thefnmark}$}\hss}}%
    \@maketitle\@thanks
  \endgroup
  \setcounter{footnote}{0}%
  \let\maketitle\relax\let\@maketitle\relax
  \gdef\@thanks{}\gdef\@author{}\gdef\@title{}\let\thanks\relax}
\renewcommand{\@maketitle}{%
  \vbox{\hsize\textwidth\linewidth\hsize
    \vskip 0.2in
    \begin{center}\LARGE\normalfont\@title\par\end{center}
    \vskip 0.04in
    \def\And{\end{tabular}\hfil\linebreak[0]\hfil
      \begin{tabular}[t]{c}\rule{\z@}{28\p@}\ignorespaces}%
    \def\AND{\end{tabular}\hfil\linebreak[4]\hfil
      \begin{tabular}[t]{c}\rule{\z@}{24\p@}\ignorespaces}%
    \begin{center}
      \begin{tabular}[t]{c}\rule{\z@}{16\p@}\@author\end{tabular}
    \end{center}
    \vskip -6.5pt}}
\makeatother
\AtBeginDocument{\raggedbottom}

\long\def\secondSplitIntroduction#1\begin#2[#3]#4\end#5#6\secondStop{%
  \ifstrequal{#2}{figure}{}{%
    \PackageError{arxiv_second}{Expected the first introduction environment to be a figure}{Check the title-page figure extraction.}}%
  \ifstrequal{#5}{figure}{}{%
    \PackageError{arxiv_second}{Unexpected nested environment in the first figure}{Check the title-page figure extraction.}}%
  \long\def\secondTitleFigure{%
    \begin{figure}[H]\captionsetup{skip=15pt}#4\end{figure}}%
  \long\def\secondIntroduction{#1#6}}
\expandafter\secondSplitIntroduction\section{Introduction}
\label{sec:introduction}

Recent advances in AI have demonstrated growing potential to accelerate scientific discovery in mathematics, biology, materials science, quantum physics, and many more~\citep{romeraparedes2024funsearch,lu2026aiscientist,gottweis2026coscientist,Wang_2023}.
Regardless of domain, existing approaches largely rely on automatic evaluations to guide the discovery process~\citep{novikov2025alphaevolve,yuksekgonul2026tttdiscover,ye2026simpletes}.
Clearly, such feedback can be unavailable or too expensive to obtain~\citep{Wang_2023}, making it difficult to assess candidate solutions and guide subsequent attempts.
While recent LLM-as-a-Judge frameworks offer a flexible way to evaluate candidates in natural language and provide feedback in the absence of task-specific automated evaluators~\citep{zheng2023llmjudge}, their judgments cannot reliably establish the validity of scientific results~\citep{tan2025judgebench}, raising the question of how automated verification can effectively support scientific discovery.

\begin{figure}[t]
\centering
\includegraphics{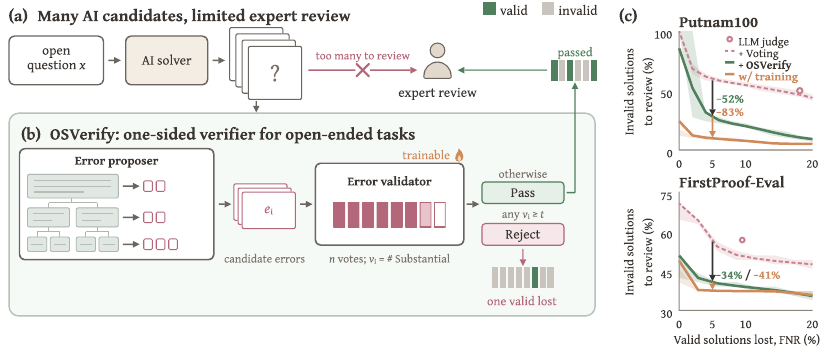}
\vspace{-10pt}
\caption{One-sided verification for open-ended scientific discovery.
(a) AI-generated candidates can outpace limited expert review capacity.
(b) \method decouples verification into hierarchical error proposal and repeated error validation, converting solution verification to error validation.
(c) Under empirical 5\% FNR constraint, \method reduces the invalid candidates requiring review by 52\% on Putnam100 and 34\% on FirstProof-Eval, and the trained variant raises the reduction to 83\% and 41\%.}
\label{fig:teaser}
\vspace{-10pt}
\end{figure}

This paper takes the view that, while AI can greatly expand the scale and breadth of future scientific exploration, human expert review remains essential to establishing the validity of its results.
Yet candidate generation can outpace experts' review capacity.
\citet{cohn2026technicaldebt} warns that producing AI-generated solutions without their verification can irresponsibly shift substantial work onto the research community.
These pressures motivate an increasingly important question: \emph{how should AI discovery systems engage human experts whose review capacity is limited?}

Verification under limited human effort should satisfy two objectives.
First, valid solutions are rare in challenging discovery problems, and mistakenly discarding one may lose an important discovery.
Second, subject to the discovery goal, a verifier should minimize the human effort spent reviewing invalid candidates.
This paper focuses on \emph{one-sided verification} under the Neyman-Pearson principle~\citep{neyman1933,tong2018npclassification}: maximize TNR subject to a low FNR constraint.
This objective requires a verifier to reject as many invalid candidates as possible while limiting the fraction of valid solutions rejected. Under a fixed FNR constraint, increasing TNR lowers the number of candidates requiring human review per valid solution found.

Drawing on the generate-and-verify intuition that checking a candidate can be easier than generating one~\citep{cobbe2021verifiers}, we hypothesize a similar difficulty gap within verification: discovering an error requires searching across a solution, whereas validating a proposed error focuses on whether that specific error invalidates the solution.
Relatedly, LLMs often fail to locate reasoning errors that they can correct once the location is given~\citep{tyen2024llms}.
Following this, this paper proposes \method, an algorithmic framework that separates verification into \emph{error proposal} and \emph{error validation}.
The error proposer searches broadly for potential errors by hierarchically decomposing the solution candidates into pieces and examining the pieces at different levels of granularity, aiming to identify errors in as many invalid solutions as possible.
The error validator checks all proposed errors and examines whether each error is minor or substantial enough to invalidate the solution.
We further post-train the error validator with GRPO on carefully constructed substantial and non-substantial proposed errors, improving its ability to distinguish different error types and generalize across problems with different difficulties.
This separation decouples the verification into two adversarial components, allowing both to improve their capability by scaling up computation effectively.

To study verification without reliable automatic evaluation, we first conduct a controlled evaluation on competition-level and research-level mathematical benchmarks with known solutions, withholding the reference solutions and formal proof checkers such as Lean from the verifier.
We intentionally use relatively weak LLMs as verifier backbones to approximate scenarios where problems exceed the models' reliable problem-solving ability. 
Under the low FNR setting, \method reduces the invalid candidates requiring human review by 46\% on average relative to the baselines. Validator training raises this reduction to 57\%, with the largest gains under the strictest constraint, and the gains carry over to research-level FirstProof-Eval, which is unseen during training.
Moreover, we demonstrate that \method can also be built upon existing verifiers and improve their one-sided performance.

Finally, as high-impact real-world use cases, we apply \method to research problems of central importance in several most challenging areas of mathematical research, including Langlands program. When reviewing 11 OpenAI preprints, \method identifies mistakes in each of these papers autonomously, which were overlooked by OpenAI internal models and might require substantial effort for human experts to discover. 
When we use \method as the verifier for a recent agentic mathematics problem solver~\citep{ju2026automatedconjectureresolutionformal}, it further helps resolve three open conjectures in relative Langlands duality~\citep{benzvi2024relativelanglandsduality,liu2026higherperiodintegralsderivatives}, whose proofs have subsequently been confirmed by human experts and are undergoing the publication process.
These results suggest that \method extends beyond controlled benchmarks and can facilitate human-AI research collaboration.

\secondStop
\AfterEndEnvironment{abstract}{\secondTitleFigure\clearpage}
\hypersetup{allcolors=osverifylink}
\DeclareCaptionFont{papercaption}{\fontsize{10}{12}\selectfont}
\title{
One-Sided Verification Under Limited Expert Review
}

\author{%
  Fan Nie\thanks{Equal contribution. Correspond to \url{{niefan,haotianye}@stanford.edu}.}\qquad
  Haotian Ye\footnotemark[1]\\[0.4em]
  Shurui Liu\qquad Zihao Wang\qquad Stefano Ermon\qquad James Zou\\[0.8em]
  Stanford University
}

\newcommand{\fnr}{\operatorname{FNR}}
\newcommand{\tnr}{\operatorname{TNR}}

\newcommand{\rej}{\textsc{Reject}\xspace}
\newcommand{\pass}{\textsc{Pass}\xspace}
\newcommand{\sol}{S}

\definecolor{auditbackground}{HTML}{F1F4EE}
\definecolor{auditborder}{HTML}{789071}
\definecolor{audittitlebackground}{HTML}{E1E8DC}
\tcbset{
  promptbox/.style={
    enhanced,
    breakable,
    colback=auditbackground,
    colframe=auditborder,
    colbacktitle=audittitlebackground,
    coltitle=osverifylink,
    fonttitle=\normalfont\bfseries,
    boxrule=0.6pt,
    arc=2mm,
    left=3mm,
    right=3mm,
    top=2.5mm,
    bottom=2.5mm,
    before skip=10pt,
    after skip=10pt
  }
}

\tcbset{
  paperpromptbox/.style={
    enhanced,
    breakable,
    listing only,
    colback=auditbackground,
    colframe=auditborder,
    coltitle=osverifylink,
    boxrule=0.55pt,
    arc=3mm,
    outer arc=3mm,
    left=3.5mm,
    right=3.5mm,
    top=5mm,
    bottom=3.5mm,
    before skip=1.1em,
    after skip=1.2em,
    drop shadow southeast={black!13!white},
    attach boxed title to top left={
      xshift=3mm,
      yshift*=-\tcboxedtitleheight/2
    },
    boxed title style={
      enhanced,
      colback=white,
      colframe=auditborder,
      boxrule=0.55pt,
      arc=2.3mm,
      outer arc=2.3mm,
      left=2.2mm,
      right=2.2mm,
      top=1.1mm,
      bottom=1.1mm
    },
    fonttitle=\normalfont\large,
    listing options={
      basicstyle=\ttfamily\small,
      breaklines=true,
      breakatwhitespace=true,
      columns=fullflexible,
      keepspaces=true,
      showstringspaces=false,
      aboveskip=0pt,
      belowskip=0pt
    }
  }
}
\let\boxedPromptInputListing\tcbinputlisting
\renewcommand{\tcbinputlisting}[1]{%
  \boxedPromptInputListing{#1,paperpromptbox}%
}

\begin{document}

\maketitle

\centerline{\normalfont\normalsize
  \href{https://osverify.github.io/}{\raisebox{-0.1ex}{\small\faGlobe}\hspace{0.4em}Website}\qquad
  \href{https://github.com/OSVerify}{\raisebox{-0.1ex}{\small\faGithub}\hspace{0.4em}GitHub}}

\begin{abstract}

AI has become increasingly helpful in scientific discovery, yet human expert review remains essential for those problems without reliable automated evaluation, posing a fundamental challenge as candidate generation scales beyond available review capacity.
This paper studies verification under limited human effort, where rare valid solutions need to be prioritized while invalid candidates should be filtered out to reduce human workload. We adopt the one-sided verification objective that maximizes the rejection rate of invalid solutions (TNR) under a constraint on the false-rejection rate of valid ones (FNR),
making human review cost an explicit term to minimize.
Drawing on the intuition that verifying a specific error can be easier than discovering one, we introduce \method, an algorithmic framework that decouples error proposal from error validation.
The error proposer searches for errors in solutions across different granularities by hierarchical decomposition, while the validator distinguishes substantial errors from minor or hallucinated ones.
Experiments on competition- and research-level benchmarks show that \method reduces the number of human reviews of invalid solutions by 46\% on average compared to voting baselines under the same empirical low FNR constraint. Our post-training of the validator further improves the reduction to 57\%. As high-impact real-world use cases, we apply \method to identify errors in published research papers that are subsequently confirmed by the original authors and help resolve three Langlands open conjectures that are confirmed by experts, demonstrating its utility to facilitate human-AI research collaboration.
\end{abstract}

\secondIntroduction
\section{Verification under Limited Human Effort}
\label{sec:formulation}

Throughout the paper, we consider verification under the LLM-as-a-Judge setting, where an LLM-based verifier $\mathcal{V}$ evaluates whether a candidate solution correctly solves a problem through reasoning, tool use, or aggregated judgments~\citep{mahdavi2025scaling,gou2024critic}. We use $(X,Y)$ to denote a random problem--solution pair and $(x,y)$ to denote a realization.
A solution $y$ is \emph{valid} if it correctly solves $x$, equivalently $y \in \sol(x)$, where $\sol(x)$ is the set of all \emph{valid} solutions for $x$. Otherwise, $y$ is \emph{invalid}. Given $(x,y) \sim (X,Y)$, a verifier $\mathcal{V}$ returns one of the decisions
\[
\mathcal{V}(x,y)\in\{\rej,\pass\},
\]
where \rej indicates that $\mathcal{V}$ judges $y \notin \sol(x)$, and \pass indicates that $\mathcal{V}$ judges $y \in \sol(x)$. Additionally, the verifier may return reasons and justifications for its decision.

Existing verification metrics do not fully reflect the requirements of human--AI collaboration in challenging, open-ended scientific discovery. Existing approaches commonly focus on accuracy, the proportion of candidates correctly passed or rejected~\citep{tan2025judgebench,mahdavi2025scaling}; however, valid solutions are typically rare in these discovery settings, so rejecting all candidates can already yield high accuracy. Alternatively, AUROC measures how well a score ranks valid and invalid candidates across thresholds~\citep{Fawcett_2006,bouchard2025uqlm}, but does not directly characterize performance in the low-FNR regime needed to preserve rare valid solutions. In addition, many scientific problems lack reliable automated evaluation~\citep{novikov2025alphaevolve,abouzaid2026firstproof}, so human experts remain essential to establishing the validity of candidates that \pass the verifier~\citep{tao2024pilot,abouzaid2026firstproof}. Neither metric directly quantifies the review effort.

\textbf{One-sided verification.}
We call for a verification objective that preserves rare valid solutions by limiting false rejections and, subject to this requirement, filters out as many invalid candidates as possible to reduce human review effort.
To this end, we adopt a Neyman--Pearson-style objective~\citep{neyman1933,tong2018npclassification} for scientific discovery, and focus on \emph{one-sided verification}: maximize the true negative rate (TNR) subject to a low false negative rate (FNR) constraint. Specifically, taking valid solutions as the positive class, \rej candidates are predicted negative and filtered out, while \pass candidates are predicted positive and retained for human review. We require the verifier $\mathcal{V}$ to satisfy, for a prescribed tolerance $\tau\in[0,1)$,
\[
\fnr(\mathcal{V})\triangleq
P(\mathcal{V}(X,Y)=\rej\mid Y \in \sol(X))
\leq\tau.
\]
Probabilities are taken over the problem--solution distribution and any randomness in the verifier. The constraint ensures that at least a $1-\tau$ fraction of valid solutions are retained by $\mathcal{V}$ for human review. To quantify the filtering of invalid candidates and its effect on human review effort, define
\[
\tnr(\mathcal{V})\triangleq
P(\mathcal{V}(X,Y)=\rej\mid Y \notin \sol(X)).
\]
The average number of \pass candidates requiring human review per retained valid solution is
\begin{equation}
\label{eq:review-ratio}
    \frac {P\big(\mathcal{V}(X,Y) = \pass\big)} {P\big(\mathcal{V}(X,Y) = \pass, Y \in \sol(X)\big)}
=   1 + \frac {(1- p) (1-\tnr(\mathcal{V}))} {p (1 - \fnr(\mathcal{V}))} \leq 1 + \frac{1-\tnr(\mathcal{V})}{p(1-\tau)},
\end{equation}
where $p \triangleq P(Y\in\sol(X))\in(0,1)$ is the proportion of valid solutions. Under the FNR constraint $\tau$, higher TNR lowers the human review cost. We therefore adopt the following objective for verification under limited human effort:
\begin{equation}
    \label{eq:obj}
    \max_{\mathcal{V}} \tnr(\mathcal{V}) \quad \text{s.t.} \quad \fnr(\mathcal{V}) \leq \tau.
\end{equation}

\begin{figure}[t]
\centering
\includegraphics[width=\textwidth]{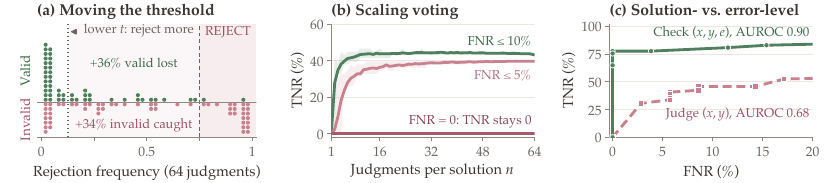}
\caption{(a) Frequencies of LLM \rej on Putnam100 with 64 samples per $(x,y)$; the dashed line marks empirical FNR below 5\%. (b) The TNR increases and saturates as the number of judgments $n$ increases, regardless of the FNR constraint. (c) LLM judgments on the error-level distribution \((x,y,e)\) achieve higher TNR at matched FNR than judgments on the solution-level distribution \((x,y)\), particularly in the low-FNR regime. All experiments use \texttt{gpt-oss-120b}.}
\vspace{-10pt}
\label{fig:motivation}
\refstepcounter{subfigure}\label{fig:motivation-scores}
\refstepcounter{subfigure}\label{fig:motivation-voting}
\refstepcounter{subfigure}\label{fig:motivation-checking}
\end{figure}

\textbf{LLM-as-a-Judge under one-sided verification.}
Contemporary LLM-based verifiers make judgments based on complex reasoning processes or agent behaviors such as tool calls and file checks. As such, it is non-trivial to obtain a numeric rejection score for the TNR--FNR trade-off.
One way to construct the score is \textsc{Voting}: sample multiple judgments independently and use the fraction of \rej votes to estimate its rejection probability~\citep{mahdavi2025scaling}. Other voting rules include pessimistic voting~\citep{huang2025pessimistic}, which rejects if any judgment returns \rej, and its optimistic counterpart, which rejects only if all judgments return \rej.

Regardless of how the rejection score is obtained, thresholding essentially selects a trade-off point on the verifier's TNR--FNR curve, rather than improving TNR under the FNR constraint.
As shown in \Cref{fig:motivation-scores}, a higher threshold preserves more valid solutions, but also passes invalid solutions whose scores fall below it. Allocating more computation to voting improves the estimate of each candidate's rejection probability, but the attainable TNR under the evaluated FNR constraints saturates in our experiment (\Cref{fig:motivation-voting}).
The central question remains unaddressed: \emph{How can we achieve a higher TNR under the same low-FNR constraint?}

\section{\method: Decoupling Error Proposal and Validation}
\label{sec:osverify-meta}
\label{sec:osverify-principle}

We build on a general intuition: checking whether a candidate satisfies a set of requirements can be easier than finding a candidate that does. For instance, finding a solution for difficult scientific problems inevitably requires searching over possible approaches, whereas evaluating a candidate focuses on a concrete attempt and is conceptually easier for the same model. This difficulty gap motivates generating multiple candidate solutions and assessing them separately~\citep{cobbe2021verifiers}, where additional compute helps scale up the search and lead to better eventual performance. We hypothesize that an analogous gap exists within verification itself. Identifying an error requires searching across a solution for what might be wrong, but given a specific proposed error, the task narrows to checking whether that particular error holds and invalidates the solution. 

Following this intuition, we introduce \method, an algorithmic framework that separates verification into error proposal, a component that searches for potential errors with broad coverage, and error validation, a component that checks whether each proposed error is substantial.
We illustrate the algorithm in \Cref{fig:teaser}. Given $(x,y)$, an \emph{error proposer} generates a set of proposed errors $\mathcal{G}(x,y)$, each $e \in \mathcal{G}(x,y)$ identifying a claim or step in $y$ and explaining why it may invalidate the solution. Its objective is high \emph{coverage}: on as many invalid solutions as possible, the proposed set should contain at least one \emph{substantial error}, i.e., a real error that invalidates the solution. An \emph{error validator} $\mathcal{C}$ then receives the full problem, solution, and one proposed error $(x,y,e)$, and checks whether the proposed error is substantial. The two stages yield the solution-level rule
\begin{equation}
\label{eq:osverify-rule}
\mathcal{V}(x,y)=\rej
\quad\Longleftrightarrow\quad
\exists e\in \mathcal{G}(x,y):\mathcal{C}\text{ confirms that }e\text{ invalidates }y.
\end{equation}

The advantage of decoupling verification into error proposal and error validation is illustrated in \Cref{fig:motivation-checking}. On FirstProof-Eval, using the same \texttt{gpt-oss-120b}, the TNR of the error-level distribution \((x,y,e)\) is higher at matched FNR than that of the solution-level distribution \((x,y)\), particularly in the low-FNR regime, with overall AUROC improving from 0.68 to 0.90. This suggests that assessing a specific proposed error is easier than determining the validity of an entire solution. These results motivate allocating additional compute to searching for potential errors, which provide concrete targets for focused validation.
More details on settings and results can be found in \Cref{sec:evaluation}.

\subsection{Error Proposer}
\label{sec:osverify-discovery}

The objective of the error proposer is to maximize the coverage of substantial errors, i.e., the probability that \(\mathcal{G}(X,Y)\) contains at least one substantial error when \(Y\notin\sol(X)\).
To this end, we search for errors at different levels of granularity within a solution, a method we call hierarchical decomposition. Inspecting the full solution exposes issues in the overall argument, while zooming in on individual parts focuses attention on specific claims and reasoning steps.

Starting from $u_1=y$ at depth zero, we recursively decompose the solution into sub-pieces. At each node $u_k$, the model selects a suitable boundary and splits the text into two contiguous parts,
\[
u_k=u_{2k}\,\Vert\,u_{2k+1},
\]
where $\Vert$ denotes concatenation in the original order. Recursion stops when the maximum depth $D$ is reached or the model judges the current part sufficiently atomic that no further split is needed. Let $U$ contain all resulting nodes, including the root, intermediate nodes, and leaves.

For each $u\in U$, the proposer receives the problem $x$, the complete solution $y$, and the part $u$ on which to focus. It returns a possibly empty set of proposed errors $\mathcal{G}(x,y,u)$, each identifying a potential error in $u$ and explaining how it could invalidate $y$. The final proposal set merges proposed errors across all nodes:
\[
\mathcal{G}(x,y)=\bigcup_{u\in U} \mathcal{G}(x,y,u).
\]
Identical proposed errors against the same target are merged, while distinct ones remain separate for validation. Algorithm~\ref{alg:error-proposer} in Appendix~\ref{sec:hierarchical-proposal} summarizes the procedure. Additionally and optionally, one may call the proposer repeatedly at selected nodes $u \in U$ to increase coverage. For instance, we find that running $\mathcal{G}(x,y,u_1)$ more than once at the root node can further improve its coverage. 

\subsection{Error Validator}
\label{sec:osverify-validation}

By maximizing the error coverage, the error proposer could also produce more hallucinated errors and minor issues that do not invalidate a solution. The objective of the error validator $\mathcal{C}$ is to assess which proposed errors warrant rejection, helping preserve valid solutions while retaining substantial errors. Specifically, given a $(x,y,e)$ tuple, it first checks whether $e$ holds, returning \textsc{Not-Real} if it does not hold, i.e., it is a hallucinated error. For a real error, it then judges whether the error invalidates $y$, returning \textsc{Substantial} if it does and \textsc{Minor} otherwise. This distinction is to account for differences in error definitions across different tasks: minor errors could lead to severe accidents in production, but are totally acceptable in proof sketches.
For each proposed error $e$, we define its substantial-error probability and the resulting solution-level rejection score as
\begin{equation}
\label{eq:validation-score}
\begin{aligned}
p_{\mathcal{C}}(x,y,e)&=P\bigl(\mathcal{C}(x,y,e)=\textsc{Substantial}\bigr),\\
s_{\mathcal{V}}(x,y)&=\max_{e\in\mathcal{G}(x,y)}p_{\mathcal{C}}(x,y,e).
\end{aligned}
\end{equation}
The maximum reflects that a single substantial error suffices to invalidate a solution, and we set the score to zero when no error is proposed. In practice, the score can be approximated by independently sampling $\mathcal{C}(x,y,e)$ multiple times.

Given a threshold $t>0$, the verifier $\mathcal V$ rejects if $s_{\mathcal{V}}(x,y)\ge t$, and returns \pass otherwise. 
For a given FNR constraint $\tau$, we select the threshold to maximize TNR subject to the constraint
\[
\fnr(\mathcal{V}_{t})=P\bigl(s_{\mathcal{V}}(X,Y)\ge t\mid Y\in\sol(X)\bigr)\le\tau.
\]
Notice that we do not consider the calibration problem here, but in practice, $t$ can be pre-estimated by a small set of samples (see example in \Cref{sec:putnam-threshold-calibration}), or by transferring from similar problem sets.
More details can be found in \Cref{sec:experiment-setup}.

\textbf{Learning to validate via RL.}
The validator operates on the same $(x,y,e)$ interface regardless of how the solution or proposed error was generated; its capability should be generalizable across different errors, solution candidates, and even problems. Additionally, the objective of the error validator is clear and binary, making it possible to construct error-level supervision and further train the validator $\mathcal C$ to distinguish substantial errors from minor or hallucinated ones across diverse problems.

Specifically, we use reinforcement learning to post-train $\mathcal C$ on the environment constructed below. 
The training data is $(x,y,e)$ and the label is whether the error $e$ is substantial or not. 
For valid problem-solution pairs $(x,y)$, all proposed errors have a negative label.
For an invalid solution $y$, since we only have the reference answer $y^\star$ and optionally the grading guideline, we provide them as the context to the proposer and the untrained validator. We sample this reference-assisted validator five times for each proposed error and label the error as positive only if at least three of the five samples judge it substantial. The remaining errors are discarded for the sake of training accuracy and informativeness.
For each training instance, we roll out the validator eight times, and use the standard GRPO loss to LoRA~\citep{hu2021loralowrankadaptationlarge} fine-tune the model. For more training details, please refer to Section~\ref{sec:experiment-setup}.

\vspace{-2pt}

\begin{table}[t]
\caption{Empirical TNR (\%) under empirical FNR constraints of 0\%, 5\%, and 10\%.
Entries report means and standard deviations over 2,000 resampling trials.
Zero includes the always-pass fallback when no nontrivial operating point satisfies the constraint.}
\label{tab:main-results}
\centering
\setlength{\tabcolsep}{1.4pt}
\footnotesize
\newcommand{\tabresult}[2]{\(\makebox[2em][r]{\(#1\)}_{\makebox[2.3em][l]{\(\scriptscriptstyle\pm #2\)}}\)}
\newcommand{\tabresultbold}[2]{{\boldmath\tabresult{#1}{#2}}}
\newcommand{\tabmissing}{\makebox[4.3em][c]{--}}
\resizebox{\textwidth}{!}{%
\begin{tabular}{lrrrrrrrrr}
\toprule
& \multicolumn{3}{c}{Putnam100} & \multicolumn{3}{c}{H2V}
& \multicolumn{3}{c}{FirstProof-Eval} \\
\cmidrule(lr){2-4}\cmidrule(lr){5-7}\cmidrule(lr){8-10}
Method  &  0\%  &  5\%  &  10\%  &  0\%  &  5\%  &  10\%  &  0\%  &  5\%  &  10\% \\
\midrule
\multicolumn{10}{c}{\texttt{gpt-oss-120b}} \\
\textsc{LLM-as-a-Judge} & \tabresult{0}{0} & \tabresult{0}{0} & \tabresult{0}{0} & \tabresult{0}{0} & \tabresult{54.4}{2.0} & \tabresult{54.4}{2.0} & \tabresult{0}{0} & \tabresult{0}{0} & \tabresult{43.3}{2.9} \\
\textsc{Voting} & \tabresult{0}{0} & \tabresult{38.4}{3.9} & \tabresult{45.5}{2.8} & \tabresult{60.9}{4.2} & \tabresult{69.2}{3.6} & \tabresult{75.6}{3.7} & \tabresult{29.5}{10.2} & \tabresult{35.9}{2.5} & \tabresult{48.8}{2.2} \\
\rowcolor{osverifygreen}
\textbf{\method}  &  \tabresultbold{0}{0}  &  \tabresultbold{70.4}{14.3}  &  \tabresultbold{77.9}{3.3}  &  \tabresultbold{83.2}{0.6}  &  \tabresultbold{84.7}{0.6}  &  \tabresultbold{84.7}{0.5}  &  \tabresultbold{50.4}{15.1}  &  \tabresultbold{58.0}{3.1}  &  \tabresultbold{60.8}{1.3} \\
\rowcolor{osverifyrose}
\  \textbf{+ trained validator} \ & \tabresultbold{85.8}{2.9} & \tabresultbold{89.6}{1.7} & \tabresultbold{92.3}{1.5} & \tabresultbold{82.4}{1.1} & \tabresultbold{83.2}{0.7} & \tabresultbold{83.2}{0.7} & \tabresultbold{57.8}{3.0} & \tabresultbold{62.3}{0.8} & \tabresultbold{62.7}{0.3} \\
\midrule
\multicolumn{10}{c}{\texttt{gpt-5-mini}} \\
\textsc{LLM-as-a-Judge} & \tabresult{0}{0} & \tabresult{0}{0} & \tabresult{63.1}{3.9} & \tabresult{66.3}{1.8} & \tabresult{66.3}{1.8} & \tabresult{66.3}{1.8} & \tabresult{0}{0} & \tabresult{0}{0} & \tabresult{42.7}{2.4} \\
\textsc{Voting} & \tabresult{28.5}{17.3} & \tabresult{58.1}{4.4} & \tabresult{75.4}{3.7} & \tabresult{82.9}{1.2} & \tabresult{82.9}{1.2} & \tabresult{82.9}{1.2} & \tabresult{19.1}{16.4} & \tabresult{42.6}{1.8} & \tabresult{49.2}{2.3} \\
\rowcolor{osverifygreen}
\textbf{\method}  &  \tabresultbold{72.1}{21.9}  &  \tabresultbold{82.9}{1.6}  &  \tabresultbold{86.6}{1.3}  &  \tabresultbold{85.7}{0.3}  &  \tabresultbold{85.7}{0.3}  &  \tabresultbold{85.7}{0.3}  &  \tabresultbold{33.9}{25.5}  &  \tabresultbold{60.0}{9.1}  &  \tabresultbold{65.4}{1.4} \\
\bottomrule
\end{tabular}}
\vspace{-10pt}
\end{table}

\section{Experiments}
\label{sec:evaluation}

Our experiments aim to address the following questions:
\textbf{(1)} How well \method can effectively reject invalid solutions while preserving valid ones, and how much RL on the error validator can further raise TNR on unseen problems;
\textbf{(2)} How verification scales up with more compute, and how \method can be applied as a meta-algorithm  on top of existing verifiers to improve their one-sided performance;
\textbf{(3)} Whether \method can support expert review in real-world research settings.

\vspace{-2pt}
\subsection{Experimental Setup}
\label{sec:experiment-setup}
\label{sec:osverify-practice}

As mentioned in \Cref{sec:introduction}, we focus on mathematical benchmarks with known solutions.
As the latest LLMs may have been too strong and could directly address these problems, we intentionally use weaker models as verifier backbones, namely \texttt{gpt-oss-120b}~\citep{agarwal2025gptoss} and \texttt{gpt-5-mini}~\citep{openai2025gpt5systemcard}, abbreviated as \oss and \mini. This simulates scenarios where problems exceed the models' problem-solving ability. Importantly, these known solutions are inaccessible to verifiers, and we do not use any formal proof checker. This allows us to study verification without reliable automated evaluation signals while rigorously assessing verifier performance.

\textbf{Datasets.}
We evaluate on three mathematical benchmarks, including competition-level Putnam100~\citep{kedlaya_putnam_archive} and Hard2Verify (H2V)~\citep{pandit2026hard2verify}, as well as research-level FirstProof-Eval constructed from First Proof~\citep{abouzaid2026firstproof} with human-labeled valid and invalid solutions.
For Putnam100, we collect publicly available official solutions as valid examples and incorrect solutions generated by frontier models. For H2V, the dataset contains model-generated solutions (valid and invalid) with step-level annotations by PhD-level mathematics experts.
Namely, both H2V and First Proof problems were released after the release of the two models considered.
For validator post-training, training data is constructed from the training portion of our collected Putnam corpus, IMO-GradingBench~\citep{luong2025imobench}, ProofBench~\citep{ma2026proofbench} and Open Proof Corpus~\citep{dekoninck2025opc}.
Details about the distribution of valid and invalid solutions and dataset construction can be found in Appendix~\ref{sec:dataset-provenance}.

\textbf{Methods.}
We compare \method against direct \textsc{LLM-as-a-Judge}~\citep{zheng2023llmjudge}, which evaluates the whole solution, as well as its \textsc{Voting} variant, which estimates the \rej probability by independently resampling the judge $n$ times. We use $n=16$ for the results in \Cref{tab:main-results} and additionally report results for $n$ ranging from $1$ to $64$ in \Cref{sec:voting-prefix64}.
We also compare with \textsc{Multiagent Debate}~\citep{du2024multiagentdebate} and pseudo-formalization with block verification (\textsc{PFBV})~\citep{barkallah2026pseudoformalization}, as specified below.
For \method, we set the maximum decomposition depth to two unless otherwise specified, with the root node at depth zero and one proposal call made at each node.
For each proposed error, we similarly estimate the validation probability with 16 independent samples.
We aggregate these judgments using the solution-level score defined in \Cref{eq:validation-score}.
For validator post-training, we use online GRPO with LoRA adapters, and keep the proposer unchanged. Appendix~\ref{app:osverify-training} describes the training procedure.

\textbf{Metrics.}
For \textsc{LLM-as-a-Judge}, we first compute the mean FNR and the mean and standard deviation of the TNR across single-call samples. At each FNR constraint, we report the corresponding TNR mean and standard deviation if the mean FNR satisfies the constraint, and report zero otherwise. For \textsc{Voting} and \method, we sweep the rejection threshold over the evaluation set and, for each empirical FNR constraint, report the highest TNR attained by any threshold satisfying that constraint, following the empirical-frontier evaluation protocol of~\citet{carlini2022membership}. The threshold used to meet each FNR constraint is determined solely from valid solutions.

\begin{figure}[t]
\centering
\includegraphics[width=\linewidth]{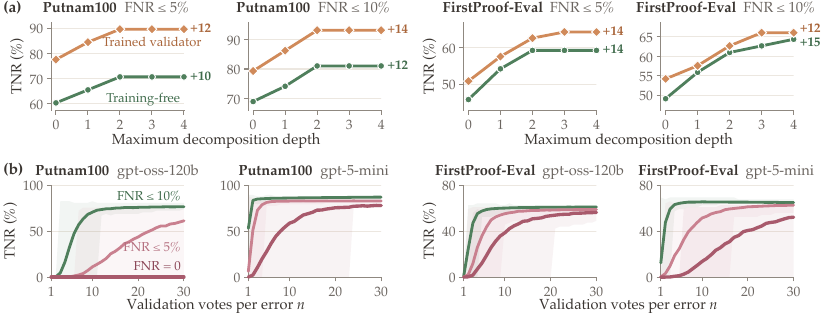}
\caption{(a) Deeper decomposition raises TNR for training-free and trained validators, with most of the gain reached by depth two.
(b) More validation votes raise TNR on fixed proposal pools; the stricter the FNR constraint, the more votes it takes.}
\label{fig:scaling}
\refstepcounter{subfigure}\label{fig:proposer-scaling}
\refstepcounter{subfigure}\label{fig:validation-scaling}
\vspace{-12pt}
\end{figure}

\subsection{\method Reduces Review Effort under the Same FNR Constraint}
\label{sec:cross-verifier-results}

Table~\ref{tab:main-results} compares \method against \textsc{LLM-as-a-Judge} and \textsc{Voting} under empirical FNR constraints of 0\%, 5\%, and 10\%.
Consistently, \method achieves a higher TNR under different FNR constraints.
With \oss, \method improves TNR over \textsc{Voting} from 38.4\% to 70.4\% on Putnam100, from 69.2\% to 84.7\% on H2V, and from 35.9\% to 58.0\% on FirstProof-Eval under the 5\% FNR constraint. 
This implies that the number of invalid candidates requiring human review is reduced by 52.0\% on Putnam100, 50.4\% on H2V, and 34.4\% on FirstProof-Eval relative to \textsc{Voting}.
With \mini, \method improves TNR over \textsc{Voting} from 58.1\% to 82.9\% on Putnam100 and from 42.6\% to 60.0\% on FirstProof-Eval under the 5\% FNR constraint.

We train the \oss-equipped error validator, with training dynamics presented in \Cref{fig:training-reward}.
Overall, the TNR can be further improved over its training-free counterpart, both on the in-distribution dataset (Putnam100) and on the out-of-distribution datasets (FirstProof-Eval).
Particularly, at zero FNR, training increases TNR from 0\% to 85.8\% on Putnam100 and from 50.4\% to 57.8\% on FirstProof-Eval. Both benchmarks retain TNR gains under the 5\% and 10\% FNR constraints. On H2V, the training-free validator already reaches the coverage limit of the proposer while rejecting few valid solutions, leaving little room for further improvement through validator training.

\subsection{Scaling Analysis and Meta-Algorithm Applications}
\label{sec:proposal-ablations}
\label{sec:verifier-extension}

Below, we investigate the scaling behavior and inference budgets of \method, and demonstrate that \method can serve as a meta-algorithm that transforms existing verifiers into one-sided verifiers.

\textbf{Scaling behavior analysis.}
Since \method can be decoupled into an error proposer and an error validator, we study how the performance of each changes as we scale up the computation cost. For the proposer, whose goal is to maximize error coverage, we increase the maximum decomposition depth and allow additional proposal calls on finer-grained parts of the solution. We use 16 validation votes per proposed error.
As shown in \Cref{fig:proposer-scaling}, under different datasets and different FNR constraints, TNR initially increases and then saturates as the maximum depth increases.
FirstProof-Eval solutions are substantially longer than Putnam100 solutions, leaving more room for decomposition. For many Putnam100 solutions, decomposition stops before reaching the depth limit.
These results support searching for error hypotheses at multiple levels of granularity.

Besides, we scale up compute for the error validator by increasing the number of resamples per error from 1 to 30, with the proposer depth fixed at 2.
As shown in \Cref{fig:validation-scaling}, the benefit of additional validation depends on the FNR constraint: performance under the 10\% constraint generally saturates with relatively few resamples, whereas tighter constraints benefit from more.
On FirstProof-Eval, increasing the vote count from 10 to 30 raises TNR under zero empirical FNR from 38.7\% to 56.5\% with \oss and from 13.6\% to 52.3\% with \mini.
Theoretically, increasing the resampling size only improves the estimation accuracy, and the performance will converge once the size is large enough.

\begin{figure}[t]
\centering
\includegraphics[width=\linewidth]{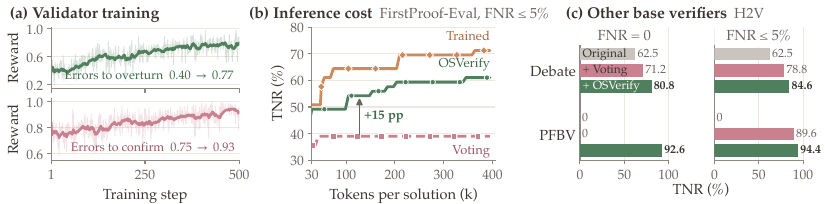}
\caption{(a) Training rewards for proposed errors to overturn or confirm, with EMA decay 0.9.
(b) Empirical TNR under different input-plus-output token budgets on FirstProof-Eval.
(c) Applying \method to existing verifiers \textsc{Multiagent Debate} and \textsc{PFBV} on H2V.}
\label{fig:training-cost-meta}
\refstepcounter{subfigure}\label{fig:training-reward}
\refstepcounter{subfigure}\label{fig:inference-cost-firstproof}
\refstepcounter{subfigure}\label{fig:base-verifier-extension}
\vspace{-20pt}
\end{figure}

\textbf{Comparison under matched token budgets.}
In \Cref{fig:inference-cost-firstproof}, we compare the performance of \textsc{Voting}, \method, and its trained counterpart under different token budgets with \oss, with more results included in Appendix~\ref{sec:cost-analysis}.
Under the 5\% FNR constraint, \method can achieve a higher TNR performance at \emph{all} token budgets per solution.
This budget includes the cost of decomposition, error proposing, and error validation.
With training, the cost is amortized and the per-token TNR further increases, and the method reaches the same TNR of 58\% with almost 4 times less compute.

\textbf{\method as a meta-algorithm.}
The proposal-validation decoupling allows \method to act as a meta-algorithm and be applied to existing verifiers that validate solutions by raising errors, either through multi-agent debate or through checking proof blocks. Specifically, for any existing verifier, we can decompose the solution into sub-pieces (if the verifier has not done this), use the verifier to find errors in the sub-pieces, and use a validator to check each error.
This allows different error-search procedures to be combined with the same validation and thresholding rule.
We apply \method to two recent verifier works, \textsc{Multiagent Debate} and \textsc{PFBV}, using \oss on H2V.
\Cref{fig:base-verifier-extension} shows that both extensions achieve higher TNR than \textsc{Voting} under the evaluated FNR constraints.
Under zero FNR, \method improves TNR over \textsc{Voting} from 71.2\% to 80.8\% for \textsc{Multiagent Debate} and from 0\% to 92.6\% for \textsc{PFBV}, and from 78.8\% to 84.6\% and from 89.6\% to 94.4\% under 5\% FNR.
These results demonstrate the applicability of \method beyond its pure LLM-based error proposer.

\subsection{Mathematical Research Case Study}
\label{sec:case-study}

We apply \method in two real-world research settings: reviewing research papers and contributing to the resolution of open conjectures. We focus on research problems of central importance in several most challenging areas of mathematical research, including Langlands program, where technically involved arguments and abstract theories require substantial specialist expertise.

As we aim to examine how \method can support research beyond the benchmarks, we turn to frontier closed-source models as the verifier backbone in these studies.

\textbf{Reviewing mathematical papers.}
We apply \method to review 11 OpenAI preprints, each of which claims to solve a central problem in an area of mathematical research. \method identifies nontrivial mathematical mistakes in each of the 11 papers autonomously, which were overlooked by OpenAI internal models and may require substantial effort for human experts to discover. These mistakes are confirmed to be real errors by human experts. The main conclusions of these papers remain valid after revisions, but correcting these issues is important to ensure the correctness of the statements and proofs and avoid confusion in the literature.
\method also discovered mistakes and suggested repairs for papers in submission to or already published in top-tier mathematical journals and induced the revisions of the authors.
These findings illustrate how \method can detect subtle mistakes in mathematical research, including previously overlooked issues in published papers and mistakes in primarily AI-generated manuscripts.

\textbf{Automated conjecture solving.}
We further pair \method with the latest open-source mathematical problem-solving system \citep{ju2026automatedconjectureresolutionformal}, a Codex agent equipped with mathematical reasoning skills, and see whether \method helps verify the candidate solutions.
No formal proof checker is used throughout the process.
We target relative Langlands duality, proposed by Ben-Zvi, Sakellaridis, and Fields medalist Venkatesh~\citep{benzvi2024relativelanglandsduality} and among the most far-reaching unifying frameworks in mathematics.
The problem is inspired by electric--magnetic duality in gauge theory, whose connection to the Langlands program was developed by Fields medalist Edward Witten and collaborators~\citep{witten2009geometriclanglandsdimensions}.
It predicts relations between periods of automorphic forms and special values of $L$-functions, unifying many earlier results~\citep[Table~1.5.1]{benzvi2024relativelanglandsduality}.
Together, the system produced proofs of three previously open statements:
(i)~the flatness of the loop-rotated Plancherel algebra $\mathrm{PL}_{X,\hbar}$ over $k[\hbar]$~\citep[Conjecture~8.5.2]{benzvi2024relativelanglandsduality};
(ii)~the commutativity of $\mathrm{PL}_X$ asserted in \citet[Corollary of Conjecture~8.1.8]{benzvi2024relativelanglandsduality}, with (i) and (ii) restated in \citet[Conjecture~4.36]{liu2026higherperiodintegralsderivatives}; and
(iii)~the automorphic commutator relation of \citet[Conjecture~4.45]{liu2026higherperiodintegralsderivatives}, which they had established only under additional cohomological vanishing assumptions and for curves of genus other than one.
\method provides a verification signal during the solving process, and returns \pass to the eventual solutions, which are examined and confirmed by human experts subsequently. The proofs of (i) and (ii) are documented in \cite{hexagon:plancherel}. The proof of (iii) is polished and rewritten in \cite{liu2026automorphiccommutatorrelationsrelative}, and it is in submission to a top-tier mathematical research journal.

Additionally, \method contributed to the paper \cite{manin2026}, which proved the Manin constant conjecture. This conjecture is of fundamental importance in number theory and the study of the BSD conjecture, and has been open for over 50 years. \method contributed in two ways: (i) it reported one mistake in an earlier version of that paper, and this mistake was overlooked by both human and 12 rounds of AI reviews using Claude Code Opus-5.5 and GPT-Astra-6 Pro (see \cite[Appendix A.3]{manin2026}); (ii) after paired with the problem-solving system \cite{ju2026automatedconjectureresolutionformal}, it reproduced the result by rejecting ineffective strategies automatically, while GPT-6-Astra Pro needed multiple rounds of prompting by mathematics experts  (see \cite[Appendix A.1, A.5]{manin2026}).
\section{Related Work}
\label{sec:related-work}

\textbf{Structured verification and error validation.}
Formal provers offer machine-checked guarantees~\citep{achim2025aristotle,chen2025seedprover}, but formalizing research proofs remains costly. In natural language, prior work splits LLM verification into smaller checks: criteria in CheckEval and LLM-as-a-Verifier~\citep{lee2025checkeval,kwok2026llmverifier}, reasoning steps in Deductive Verification~\citep{ling2023deductive}, Pessimistic Verification~\citep{huang2025pessimistic}, and self-contained proof blocks in PFBV~\citep{barkallah2026pseudoformalization}.
Judging such units in isolation can produce spurious failures when context is lost~\citep{wanner2024closer,hu2025decomposition}, so \method keeps the full problem and solution in context at every level of its decomposition.
DeepSeekMath-V2 trains a verifier to identify issues in math proofs~\citep{shao2025deepseekmathv2}.
\method combines error check with decomposition: it proposes errors at every level and rejects a solution only when independent validation confirms one of them.

\textbf{One-sided objectives.}
LLM-based checkers already find errors in published papers that experts confirm, but with low precision~\citep{son2025spot}, which motivates an explicit limit on false rejections. Our objective follows the Neyman--Pearson paradigm, which constrains one class-conditional error rate while optimizing the other~\citep{neyman1933,scott2005neyman,tong2018npclassification}.
FactTest~\citep{pmlr-v267-nie25a} applies this principle to factuality with finite-sample guarantees; \Cref{sec:putnam-threshold-calibration} uses its threshold selection on held-out valid solutions.
\Cref{app:additional-related-work} discusses additional related work about AI for discovery, LLM judges, and learning verifiers.

\section{Conclusion}

We study scientific verification under limited human review capacity, aiming to filter out invalid solutions while preserving valid candidates for expert review.
We introduce \method, which separates error proposal from error validation to improve TNR under low FNR constraints.
Experiments on three mathematical benchmarks demonstrate its effectiveness, with further gains from validator training.
Applications to mathematical paper review and open conjecture solving yield findings confirmed by human experts.
These results highlight the potential of structured verification to support scientific discovery with limited human effort.

\section*{Acknowledgments}
We thank Ken Ziyu Liu and Yongchan Kwon for helpful discussions, advice, and support.
We also thank members of the Zou group for insightful discussions.

\clearpage

\bibliography{references}
\bibliographystyle{plainnat}

\clearpage
\appendix
\section*{Appendix}
\phantomsection
\addcontentsline{toc}{part}{Appendix}
\etocsettocstyle{\subsection*{Contents}}{}
\etocsetnexttocdepth{subsection}
\localtableofcontents
\clearpage

\Needspace{6\baselineskip}
\section{Additional Related Work}
\label{app:additional-related-work}
This section extends Section~\ref{sec:related-work} with broader context on AI-driven discovery, LLM judges, and learning verifiers.

\paragraph{AI for scientific discovery and expert review.}
AI systems increasingly support scientific research, from hypothesis generation and experimentation~\citep{Wang_2023,boiko2023emergent} to research agents such as The AI Scientist, Virtual Lab, and AI Co-Scientist~\citep{lu2026aiscientist,swanson2025virtuallab,gottweis2026coscientist}.
Automated evaluators drive large-scale search in FunSearch, AlphaEvolve, TTT-Discover, and SimpleTES~\citep{romeraparedes2024funsearch,novikov2025alphaevolve,yuksekgonul2026tttdiscover,ye2026simpletes}; in natural-language mathematics, Aletheia interleaves generation, verification, and revision~\citep{feng2026aletheia}, and recent reports document model-assisted progress on research problems~\citep{feng2026erdos,woodruff2026gemini}.
Critique-assisted evaluation, such as self-critiquing models, CriticGPT, and Critique-Resilient Benchmarking, helps humans find flaws while keeping them as final arbiters~\citep{saunders2022selfcritiquing,mcaleese2024critics,marro2026benchmarking}.
We study the verifier's side of this division of labor: filtering candidates for limited expert review without discarding rare valid solutions.

\paragraph{LLM judges, self-correction, and aggregation.}
\citet{Gu_2026} survey LLM-as-a-Judge methods and strategies for improving their reliability, consistency, and robustness to bias.
\textsc{LLM-as-a-Judge} and G-Eval use language models to evaluate generated outputs~\citep{zheng2023llmjudge,liu2023geval}, and JudgeBench and CriticEval examine the reliability of such judgments and critiques~\citep{tan2025judgebench,lan2024criticeval}.
Self-correction without external feedback can fail to improve reasoning~\citep{huang2023selfcorrect}, and LLMs often fail to locate reasoning errors that they can correct once the location is given~\citep{tyen2024llms}.
Self-Refine and CRITIC revise outputs using critiques, with CRITIC incorporating tool feedback~\citep{madaan2023selfrefine,gou2024critic}.
Multiagent Debate refines answers through exchanges among model instances~\citep{du2024multiagentdebate,khan2024debating}, Multi-Agent Verification aggregates binary judgments from multiple aspect verifiers~\citep{lifshitz2025mav}, and prompt ensembles improve proof verification with smaller models~\citep{naik2026frontier}.

\paragraph{Learning verifiers.}
Outcome and process verifiers learn from solution-level and step-level supervision, respectively~\citep{cobbe2021verifiers,lightman2024verify}, and Generative Verifiers formulate reward modeling as next-token prediction~\citep{zhang2025generativeverifiers}.
Math-Minos incorporates natural-language feedback into verifier training~\citep{gao2025mathminos}, reinforcement learning has been used to train LLM judges and critics~\citep{whitehouse2026j1,xie2025ctrl}, and \citet{mahdavi2025scaling} study how inference compute and reinforcement learning affect proof verification.
These approaches supervise judgments of whole solutions, steps, or critiques; we instead train a validator on individual proposed errors, predicting whether each one warrants rejection.

\Needspace{6\baselineskip}
\section{Additional Results and Analyses}
\label{sec:additional-results}

\subsection{Additional training-free comparisons}

Table~\ref{tab:full-training-free-results} extends the training-free rows of Table~\ref{tab:main-results} with two fixed voting rules.
Pessimistic voting~\citep{huang2025pessimistic} rejects when any of the 16 sampled judgments rejects the solution.
Optimistic voting rejects only when all 16 judgments reject it.
We recompute both rules over 2,000 draws from the recorded pools.
Within each draw, a rule that violates the FNR constraint is replaced by the always-pass rule before averaging TNR.

\begin{table}[H]
\caption{Additional training-free comparisons. Empirical TNR (\%) under empirical FNR constraints of 0\%, 5\%, and 10\%, reported as mean and standard deviation over 2,000 resampling trials.
\textsc{LLM-as-a-Judge}, \textsc{Voting}, and \method entries are reproduced from Table~\ref{tab:main-results}.
Pessimistic and optimistic voting use 16 judgments drawn without replacement from each recorded 64-judgment pool; an infeasible rule is replaced by the always-pass rule within each trial.}
\label{tab:full-training-free-results}
\centering
\setlength{\tabcolsep}{1.4pt}
\footnotesize
\newcommand{\tabresult}[2]{\(\makebox[2em][r]{\(#1\)}_{\makebox[2.3em][l]{\(\scriptscriptstyle\pm #2\)}}\)}
\newcommand{\tabresultbold}[2]{{\boldmath\tabresult{#1}{#2}}}
\newcommand{\tabmissing}{\makebox[4.3em][c]{--}}
\resizebox{\textwidth}{!}{%
\begin{tabular}{lrrrrrrrrr}
\toprule
& \multicolumn{3}{c}{Putnam100} & \multicolumn{3}{c}{H2V}
& \multicolumn{3}{c}{FirstProof-Eval} \\
\cmidrule(lr){2-4}\cmidrule(lr){5-7}\cmidrule(lr){8-10}
Method  &  0\%  &  5\%  &  10\%  &  0\%  &  5\%  &  10\%  &  0\%  &  5\%  &  10\% \\
\midrule
\multicolumn{10}{c}{\texttt{gpt-oss-120b}} \\
\textsc{LLM-as-a-Judge} & \tabresult{0}{0} & \tabresult{0}{0} & \tabresult{0}{0} & \tabresult{0}{0} & \tabresult{54.4}{2.0} & \tabresult{54.4}{2.0} & \tabresult{0}{0} & \tabresult{0}{0} & \tabresult{43.3}{2.9} \\
Pessimistic voting & \tabresult{0}{0} & \tabresult{0}{0} & \tabresult{0}{0} & \tabresult{0}{0} & \tabresult{1.8}{11.8} & \tabresult{45.6}{38.8} & \tabresult{0}{0} & \tabresult{0}{0} & \tabresult{0}{0} \\
Optimistic voting & \tabresult{0}{0} & \tabresult{19.7}{2.6} & \tabresult{19.7}{2.6} & \tabresult{33.3}{1.4} & \tabresult{33.3}{1.4} & \tabresult{33.3}{1.4} & \tabresult{26.8}{9.1} & \tabresult{29.8}{1.9} & \tabresult{29.8}{1.7} \\
\textsc{Voting} & \tabresult{0.0}{0.0} & \tabresult{38.4}{3.9} & \tabresult{45.5}{2.8} & \tabresult{60.9}{4.2} & \tabresult{69.2}{3.6} & \tabresult{75.6}{3.7} & \tabresult{29.5}{10.2} & \tabresult{35.9}{2.5} & \tabresult{48.8}{2.2} \\
\rowcolor{osverifygreen}
\textbf{\method}  &  \tabresultbold{0}{0}  &  \tabresultbold{70.4}{14.3}  &  \tabresultbold{77.9}{3.3}  &  \tabresultbold{83.2}{0.6}  &  \tabresultbold{84.7}{0.6}  &  \tabresultbold{84.7}{0.5}  &  \tabresultbold{50.4}{15.1}  &  \tabresultbold{58.0}{3.1}  &  \tabresultbold{60.8}{1.3} \\
\midrule
\multicolumn{10}{c}{\texttt{gpt-5-mini}} \\
\textsc{LLM-as-a-Judge} & \tabresult{0}{0} & \tabresult{0}{0} & \tabresult{63.1}{3.9} & \tabresult{66.3}{1.8} & \tabresult{66.3}{1.8} & \tabresult{66.3}{1.8} & \tabresult{0}{0} & \tabresult{0}{0} & \tabresult{42.7}{2.4} \\
Pessimistic voting & \tabresult{0}{0} & \tabresult{0}{0} & \tabresult{0}{0} & \tabresult{82.9}{1.2} & \tabresult{82.9}{1.2} & \tabresult{82.9}{1.2} & \tabresult{0}{0} & \tabresult{0}{0} & \tabresult{0}{0} \\
Optimistic voting & \tabresult{25.7}{15.2} & \tabresult{34.6}{2.4} & \tabresult{34.6}{2.4} & \tabresult{50.9}{1.2} & \tabresult{50.9}{1.2} & \tabresult{50.9}{1.2} & \tabresult{18.7}{16.0} & \tabresult{32.3}{1.5} & \tabresult{32.3}{1.5} \\
\textsc{Voting} & \tabresult{28.5}{17.3} & \tabresult{58.1}{4.4} & \tabresult{75.4}{3.7} & \tabresult{82.9}{1.2} & \tabresult{82.9}{1.2} & \tabresult{82.9}{1.2} & \tabresult{19.1}{16.4} & \tabresult{42.6}{1.8} & \tabresult{49.2}{2.3} \\
\rowcolor{osverifygreen}
\textbf{\method}  &  \tabresultbold{72.1}{21.9}  &  \tabresultbold{82.9}{1.6}  &  \tabresultbold{86.6}{1.3}  &  \tabresultbold{85.7}{0.3}  &  \tabresultbold{85.7}{0.3}  &  \tabresultbold{85.7}{0.3}  &  \tabresultbold{33.9}{25.5}  &  \tabresultbold{60.0}{9.1}  &  \tabresultbold{65.4}{1.4} \\
\bottomrule
\end{tabular}}
\end{table}

\subsection{Cost analysis}
\label{sec:cost-analysis}

We compare \textsc{Voting}, \method, and \method with a trained validator under approximate token budgets.
Cost is the mean number of input and output tokens per solution, including reasoning tokens, summed over all stages: judgments for \textsc{Voting}; decomposition, error proposal, and validation for \method.
We estimate token counts from recorded usage and saved text; decomposition cost grows with the selected depth, and validation cost with the number of votes.
This comparison uses tokens as a proxy for inference compute.

\textsc{Voting} scales the number of judgments per solution from 1 to 64.
For both \method configurations, we jointly vary the number of validation votes from 1 to 16 and the decomposition depth from 0 to 4.
Each point selects the highest TNR within the token budget under the indicated empirical FNR constraint.

\begin{figure}[H]
\centering
\includegraphics[width=\linewidth]{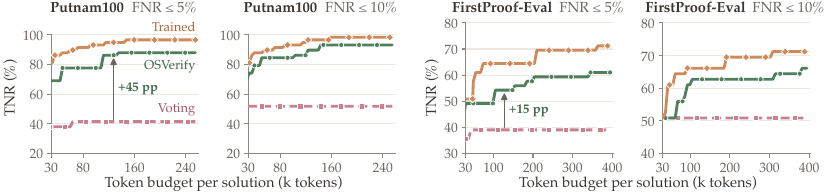}
\caption{\textbf{Approximate inference cost comparison.}
\method never falls below \textsc{Voting} at any displayed budget; arrows mark its gain at 128k tokens per solution.
Each point reports the best empirical TNR attainable within an input-plus-output token budget.
\textsc{Voting} varies from 1 to 64 judgments; \method jointly varies the proposal budget and 1--16 validation votes per proposed error.
The displayed range starts at 30k tokens per solution, and curves omit leading zero-TNR operating points that accept every solution.
Lines connect evaluated budget points; vertical axes vary across panels.}
\label{fig:token-budget-cost-analysis}
\end{figure}

Figure~\ref{fig:token-budget-cost-analysis} shows that \method attains higher TNR than \textsc{Voting} over the displayed budget range.
The trained-validator configuration further improves the attainable frontier on Putnam100 and FirstProof-Eval.
At approximately 128k tokens per solution and a 5\% FNR constraint on Putnam100, \textsc{Voting}, \method, and the trained-validator configuration attain TNRs of 41.4\%, 86.2\%, and 94.8\%, respectively.

\paragraph{Component-wise inference cost.}
Table~\ref{tab:h2v-cost-accounting} reports the decomposition, proposal, and validation costs for 200 H2V solutions with \oss, decomposition depth two, and 16 validator judgments per proposed error.
Decomposition and proposer costs use recorded provider usage.
Validator input tokens are reconstructed with the model tokenizer, and output tokens are estimated from the mean length of 50 previously sampled completions.
The estimate counts the prompt for each proposed error once and sums the output tokens of 16 completions under batched-equivalent accounting.

\begin{table}[H]
\caption{Component-wise H2V inference cost with \oss, projected to 16 validator judgments per proposed error.
Decomposition and proposer usage are provider-reported; validator inputs are reconstructed with the model tokenizer.
Outputs marked \(\dagger\) are estimated from previously sampled completion lengths.
Requests include every internal decomposition request and one batched-equivalent validation request per proposed error.
\(K=10^3\) and \(M=10^6\).}
\label{tab:h2v-cost-accounting}
\centering
\setlength{\tabcolsep}{5pt}
\resizebox{\textwidth}{!}{%
\begin{tabular}{lrrrrr}
\toprule
Component & Requests & Input tokens & Output tokens & Total tokens & Tokens/solution \\
\midrule
\method: decomposition
& 4.51K & 7.08M & 3.26M & 10.34M & 51.7K \\
\method: proposer
& 1.19K & 6.47M & 2.28M & 8.75M & 43.7K \\
\method: validator (\(n=16\))
& 672 & 3.11M & \(15.22\mathrm{M}^\dagger\)
& \(18.33\mathrm{M}^\dagger\) & 91.6K \\
\bottomrule
\end{tabular}}
\end{table}

The validation estimate uses the original proposed errors and prompts, with completion lengths extrapolated to 16 judgments per proposed error.
All 50 sampled validator completions terminated without reaching the output-token limit.

\subsection{Saturation of the \textsc{Voting} baseline}
\label{sec:voting-prefix64}

\begin{figure}[H]
\centering
\includegraphics[width=\linewidth]{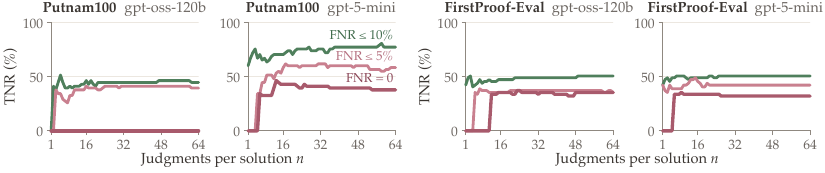}
\caption{\textbf{Scaling the whole-solution \textsc{Voting} baseline to 64 judgments.}
Each curve reports the best empirical TNR at the FNR constraint of 0\%, 5\%, or 10\%, using the first $n$ judgments per solution. No subset resampling or smoothing is applied.}
\label{fig:voting-prefix64}
\end{figure}

Figure~\ref{fig:voting-prefix64} scales the whole-solution \textsc{Voting} baseline from 1 to 64 judgments for two backbones on Putnam100 and FirstProof-Eval.
We use the first $n$ recorded judgments for each solution, without subset resampling.
At each $n$, we select the rejection threshold with the highest TNR under the specified empirical FNR constraint on the evaluated cohort.

\textsc{Voting} exhibits diminishing returns, with most curves approaching a plateau within the first few dozen judgments.
For \oss under the 5\% FNR cap, increasing the vote count from 16 to 64 changes TNR from 41.4\% to 39.7\% on Putnam100 and leaves it at 35.6\% on FirstProof-Eval.
Thus increasing \textsc{Voting} to 64 judgments does not recover the main-table gains from \method on these configurations.
These are empirical observations within the evaluated budget range.

\subsection{Held-out threshold selection on Putnam}
\label{sec:putnam-threshold-calibration}

The main results select thresholds on the evaluation set; this section illustrates deployment, where the threshold is instead selected on held-out valid solutions.
We use 100 Putnam valid solutions from the training pool as a calibration dataset. They share no problems with the Putnam100 solutions.
We collect 16 whole-solution judgments for \textsc{Voting} and 16 validation judgments per proposed error for \method, using \oss.
The score is the number of rejecting votes for \textsc{Voting} and the maximum substantial-vote count over proposed errors for \method.

We select thresholds using the order-statistic procedure of FactTest~\citep{pmlr-v267-nie25a}, with valid solutions as the null class and $\delta=0.10$.
For a target valid-solution rejection rate $\tau$, we choose the smallest $k$ satisfying
\[
\sum_{j=k}^{100}\binom{100}{j}(1-\tau)^j\tau^{100-j}\leq\delta.
\]
We set the threshold to the $k$th smallest calibration score and reject only scores strictly above it.
This gives $k=99$ for $\tau=0.05$ and $k=95$ for $\tau=0.10$.
Table~\ref{tab:putnam-threshold-calibration} applies the resulting thresholds to the recorded test votes.

\begin{table}[H]
\centering
\caption{\textbf{Diagnostic comparison using calibration-selected thresholds on Putnam.}
All scores use 16 votes.}
\label{tab:putnam-threshold-calibration}
\setlength{\tabcolsep}{6pt}
\begin{tabular}{clccc}
\toprule
Target FNR & Method & Reject if score & Test FNR & Test TNR \\
\midrule
5\% & \textsc{Voting} & $>16$ & 0.0\% (0/50) & 0.0\% (0/58) \\
& \method & $\geq16$ & 4.0\% (2/50) & 77.6\% (45/58) \\
\midrule
10\% & \textsc{Voting} & $\geq15$ & 2.0\% (1/50) & 29.3\% (17/58) \\
& \method & $\geq15$ & 6.0\% (3/50) & 81.0\% (47/58) \\
\bottomrule
\end{tabular}
\end{table}

\subsection{Training data audit ablation}
\label{sec:training-audit-ablation}

\begin{figure}[H]
\centering
\includegraphics[width=0.62\linewidth]{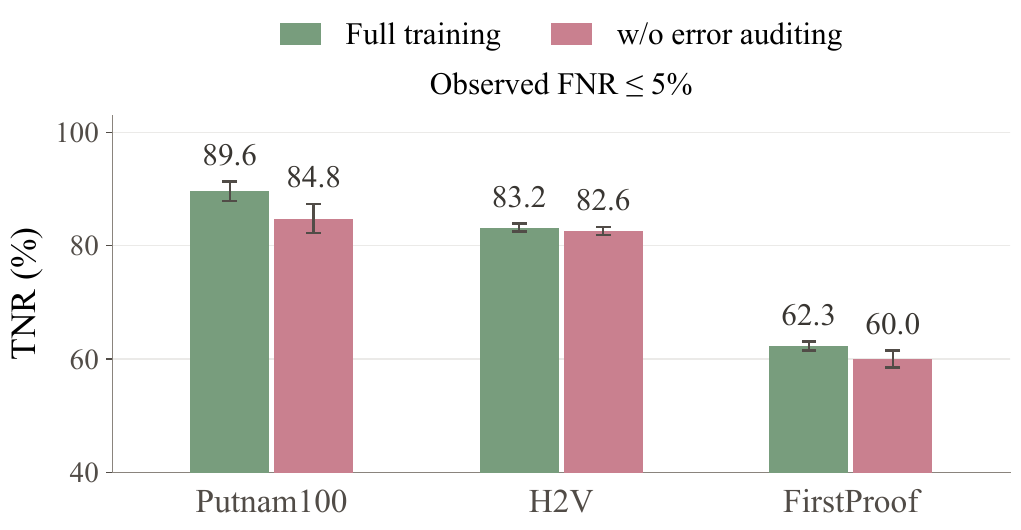}
\caption{\textbf{Auditing proposed errors for validator training.}
We compare the full training setup with a variant that omits auditing proposed errors from invalid solutions.
Bars show mean TNR under a 5\% empirical FNR constraint; error bars show standard deviations.
The vertical axis starts at 40\%.}
\label{fig:training-audit-ablation}
\end{figure}

We examine the effect of auditing proposed errors from invalid solutions when constructing validator training data.
The full setup uses the reference-assisted screening described in Appendix~\ref{app:training-data}; the ablated setup omits this audit.
Figure~\ref{fig:training-audit-ablation} compares the resulting \oss validators under a 5\% empirical FNR constraint.
With auditing, TNR is $89.6\pm1.7\%$ on Putnam100, $83.2\pm0.7\%$ on H2V, and $62.3\pm0.8\%$ on FirstProof-Eval.
Without auditing, the corresponding results are $84.8\pm2.6\%$, $82.6\pm0.7\%$, and $60.0\pm1.5\%$.
Auditing improves mean TNR by 4.8, 0.6, and 2.3 percentage points, respectively.
These results support checking individual proposed errors rather than relying solely on the invalidity of the source solution.

\Needspace{6\baselineskip}
\section{Datasets and Annotation}
\label{app:dataset}
\label{sec:dataset-provenance}

\subsection{Evaluation datasets}
\label{app:evaluation-datasets}

Table~\ref{tab:dataset-cohorts} summarizes the evaluation datasets. Valid
solutions are the positive class for the solution-level FNR and TNR
definitions.

\begin{table}[H]
\caption{Evaluation datasets.}
\label{tab:dataset-cohorts}
\centering
\small
\begin{tabular}{lrrr}
\toprule
Benchmark & Invalid & Valid & Total \\
\midrule
Putnam100 & 58 & 50 & 108 \\
H2V & 163 & 37 & 200 \\
FirstProof-Eval & 59 & 35 & 94 \\
\bottomrule
\end{tabular}
\end{table}

\paragraph{Putnam100.}
The valid set contains 50 official solutions to 46 Putnam problems, collected from the Putnam archive.\footnote{\url{https://kskedlaya.org/putnam-archive/}}
To construct the invalid set, we generate solutions using Claude Sonnet 4.5, GPT-5.4-mini, and Qwen3-235B-A22B-Instruct.
We grade these solutions against the official references using GPT-5.5 with xhigh reasoning effort and retain those classified as invalid.
Human experts then independently verify that the retained solutions are indeed invalid.

\paragraph{H2V.}
We use the full test split of Hard2Verify~\citep{pandit2026hard2verify}, which contains 200 solutions to 79 problems, generated by GPT-5, Claude Sonnet 4, and Gemini 2.5 Pro, with 1,860 step-level labels from expert annotators. A solution is valid if and only if all of its steps are labeled correct.
A solution is valid if and only if all of its steps are labeled correct, which yields 42 valid and 158 invalid solutions. 
We subsequently re-examine all valid and invalid solutions. We relabel five solutions from valid to invalid because they contain a false step or omits a necessary component of the argument.
The relabeling is independently confirmed by a human expert.
Among them, four contain a false step: a circular equivalence between an event and itself (solution 140), an appeal to a nonexistent theorem whose claimed probability of \(1/2\) contradicts the official value of about \(1/3\) (141), an upper bound silently used as an equality (199), and a false
structural lemma (62; see the example below). The fifth (179) omits the upper-bound direction of the proof, which the official solution establishes over a full page.

\paragraph{FirstProof-Eval.}
FirstProof-Eval contains solutions to the 20 problems of First Proof~\citep{abouzaid2026firstproof}, released in two batches of ten. The 94 solutions comprise 17 written by humans, 39 generated by GPT and Gemini models, and 38 released publicly by OpenAI, UCLA, Princeton, and IMProofBench, each with a human-assigned validity label. 

\begin{tcolorbox}[promptbox, title={Audit example: H2V solution 62}]
\small
\textbf{Problem.} A rectangle \(R\) is divided into finitely many
axis-parallel rectangles, no three of which share a corner. An ant moves
from a corner of a chosen rectangle to an adjacent corner of the same
rectangle. Prove that if the ant travels from the bottom-left to the
top-right corner of \(R\), some rectangle is chosen at least twice.

\textbf{Original label.} Valid.

\textbf{Claim in the solution (Step 0).} Every interior cut runs across
all of \(R\), so the dissection consists of parallel strips. The argument
considers an interior endpoint \(P\) of a vertical cut and asserts that
``the rectangle(s) adjacent above or below then also have a corner at
\(P\),'' which would give three rectangles with a common corner.

\textbf{Counterexample.} Let \(R=[0,2]^2\), \(A=[0,2]\times[1,2]\),
\(B=[0,1]\times[0,1]\), and \(C=[1,2]\times[0,1]\). The vertical cut
\(x=1\) ends at \(P=(1,1)\), which is a corner of \(B\) and \(C\) only;
\(P\) lies in the interior of the bottom side of \(A\). No point is a
corner of three rectangles, yet the dissection is not a strip
decomposition.

\textbf{Consequence.} Steps~2--3 count moves under the strip structure, so the proof does not establish the result, and the step is not locally repairable. We relabel the solution as invalid.
\end{tcolorbox}

\subsection{Training data and supervision}
\label{app:training-data}
We construct training examples from the training split of the Putnam corpus, IMO-GradingBench~\citep{luong2025imobench}, ProofBench~\citep{ma2026proofbench}, and the Open Proof Corpus~\citep{dekoninck2025opc}.
Each example consists of a problem, a submitted solution, a proposed error, and a binary target indicating whether the error warrants rejection.
For valid source solutions, proposed errors receive negative targets.
For invalid source solutions, trusted reference answers and available grading guidelines support error proposal and screening before a positive target is assigned; only proposed errors judged substantial by at least 3 of the 5 validation votes receive positive targets.
The online validator receives the reference-free problem, solution, and proposed error.
We exclude easy examples, on which all five sampled verdicts of the base validator match the target, and examples whose prompts leave fewer than 8,192 output tokens in the 32,768-token context window.
Table~\ref{tab:training-pool} gives the resulting composition.

\begin{table}[H]
\caption{Composition of the online GRPO training pool. Negative examples are proposed errors on valid solutions; positive examples are substantial proposed errors on invalid solutions.}
\label{tab:training-pool}
\centering
\small
\begin{tabular}{lrrr}
\toprule
Source & Negative & Positive & Total \\
\midrule
Putnam (training split) & 774 & 270 & 1,044 \\
IMO-GradingBench & 71 & 165 & 236 \\
ProofBench & 45 & 223 & 268 \\
Open Proof Corpus & 13 & 126 & 139 \\
\midrule
Total & 903 & 784 & 1,687 \\
\bottomrule
\end{tabular}
\end{table}

\Needspace{6\baselineskip}
\section{Experimental Details}
\label{sec:implementation-details}

\subsection{Evaluation protocol and metrics}
\label{sec:metrics-uncertainty}

For Voting and \method, we independently sample judgments without replacement from each recorded pool.
Within each trial, we reselect the rejection threshold that maximizes empirical TNR under the specified FNR constraint, including the always-pass rule.

The reported means and standard deviations summarize variation across the 2,000 resampling trials with the evaluation solutions held fixed.

\subsection{Models and inference configurations}
\label{sec:model-configs}
Table~\ref{tab:model-configs} lists the backbones and their inference settings.
The main comparison uses the sampling protocol above.
Unless otherwise specified, we use the same backbone for the error proposer and error validator in \method.
All validation calls include the problem, the full submitted solution, the focused excerpt, and the proposed error.
Dataset-specific policies are fixed within each experiment.

\begin{table}[H]
\caption{Backbones and inference settings. \mini only supports
its default temperature of 1. The trained validator is \oss
with the LoRA adapter described in Appendix~\ref{app:osverify-training}; it
validates the proposed errors of the training-free \oss
proposer.}
\label{tab:model-configs}
\centering
\small
\begin{tabular}{llcc}
\toprule
Model & Provider & Reasoning effort & Temperature \\
\midrule
\oss & Together AI & Medium & 0.7 \\
\mini & OpenAI & Medium & 1.0 \\
\oss + LoRA (trained validator) & Tinker & Medium & 0.7 \\
\bottomrule
\end{tabular}
\end{table}

\subsection{Setup of whole-solution judging and proposed-error checking}
\label{sec:judge-vs-check-details}

\Cref{fig:motivation-checking} compares \oss judgments of whole solutions with its validation of proposed errors on FirstProof-Eval.
Both use 15 votes per decision. For whole-solution judging, the score is the fraction of \textsc{Reject} verdicts among the first 15 whole-solution judgments. At each threshold, TNR is the fraction of the 59 invalid solutions rejected, and FNR is the fraction of the 35 valid solutions rejected.

For proposed-error checking, the score is the fraction of 15 validator votes that label the proposed error \textsc{Substantial}. 
TNR is the fraction of the errors on that are confirmed as \textsc{Substantial} by an independent audit (gpt-6-astra xhigh), and FNR is the fraction of the errors on that are confirmed as \textsc{Minor} or \textsc{Not-real} by the audit.

\subsection{Hierarchical proposal and aggregation}
\label{sec:hierarchical-proposal}
We use maximum decomposition depth two for the main comparison, with the root at depth zero and one proposal call per retained node.
The solution-level score is the largest number of \textsc{Substantial} votes received by any of its proposed errors; solutions without proposed errors receive score zero.
The decision rejects if this score reaches the selected positive threshold.

\begin{algorithm}[H]
\caption{Hierarchical error proposal}
\label{alg:error-proposer}
\begin{minipage}{\linewidth}
\textbf{Input}\quad Problem $x$, solution $y$, maximum depth $D$.

\textbf{Function} $\operatorname{Decompose}(u_k,d)$:
\begin{enumerate}[leftmargin=1.7em,itemsep=0.1em,topsep=0.3em]
\item If $d=D$ or the model judges $u_k$ atomic, return $\{u_k\}$.
\item Split $u_k$ into $(u_{2k},u_{2k+1})$; return $\{u_k\}\cup\operatorname{Decompose}(u_{2k},d+1)\cup\operatorname{Decompose}(u_{2k+1},d+1)$.
\end{enumerate}
\textbf{Main}:
\begin{enumerate}[start=3,leftmargin=1.7em,itemsep=0.1em,topsep=0.3em]
\item Set $u_1=y$, $U\gets\operatorname{Decompose}(u_1,0)$ and $\mathcal{G}(x,y)\gets \bigcup _{u \in U} \mathcal{G}(x,y,u)$.
\item Merge identical proposed errors and return $\mathcal{G}(x,y)$.
\end{enumerate}
\end{minipage}
\end{algorithm}

\subsection{Online GRPO training}
\label{app:osverify-training}

We fine-tune the \oss validator with LoRA using online GRPO~\citep{shao2024deepseekmath}, starting from the base model.
Training inputs and targets are constructed as described in Appendix~\ref{app:training-data}.
At each optimization step, we sample 32 training examples and the current policy generates eight fresh rollouts for each.
Each rollout uses the validator prompt with the dataset policy of its source domain.
Examples are sampled with weights that balance valid-side and invalid-side examples within each domain, and the oversampling of any single example is capped at three times that of the least-sampled example in its domain.
Each rollout receives reward one if its final verdict matches the binary target and zero otherwise.
For positive targets, the required verdict is \textsc{Substantial}; for negative targets, either \textsc{Not-Real} or \textsc{Minor} is accepted.
Truncated responses and responses without a parseable final verdict receive zero reward.

For rewards $r_{i1},\ldots,r_{i8}$ from example $i$, we compute the within-group mean $\bar r_i$ and standard deviation $\sigma_i$.
The normalized advantage is
\[
A_{ij}=\frac{r_{ij}-\bar r_i}{\sigma_i+\epsilon}.
\]
Groups with zero reward variance do not contribute an update.
A per-token KL penalty against the base model is subtracted from the advantage of each generated token.
We take one policy-gradient update on the fresh rollouts per step, summing contributions over generated tokens without per-sequence length normalization, and then refresh the sampling policy.
We report the checkpoint after 400 optimization steps.
Table~\ref{tab:grpo-hyperparameters} lists the hyperparameters.

\begin{table}[H]
\caption{Online GRPO hyperparameters.}
\label{tab:grpo-hyperparameters}
\centering
\small
\begin{tabular}{ll}
\toprule
Setting & Value \\
\midrule
Base model & \oss, medium reasoning effort \\
Adapter & LoRA, rank 32 \\
Optimizer & Adam, $\beta_1=0.9$, $\beta_2=0.95$, $\epsilon=10^{-8}$ \\
Learning rate & $10^{-5}$, linear warmup for 10 steps, then constant \\
Examples per step & 32 \\
Rollouts per example & 8 \\
Sampling temperature & 0.7 \\
Context window & 32,768 tokens (prompt and completion) \\
KL coefficient & 0.03 (reference: base model) \\
Advantage $\epsilon$ & $10^{-6}$ \\
Per-example oversampling cap & $3\times$ \\
Training Steps & 400 steps \\
\bottomrule
\end{tabular}
\end{table}

\Needspace{6\baselineskip}
\section{Prompts}
\label{sec:prompts}

The listings below materialize the prompt code used for the reported
training-free experiments. Bracketed fields are replaced at inference time.
We normalize typographic punctuation to ASCII for display; the instructions
and output schemas are otherwise unchanged.

\subsection{Proposer prompt}

The box below gives the root-level proposer prompt. For a non-root node, the
problem and full solution remain visible, the node's
verbatim solution excerpt is added, and root-level
\texttt{TASK\_INCOMPLETE} errors are disabled to avoid decomposition
boundary artifacts. The rules for proposed errors, the error types, and the
JSON schema are shared across depths.

\tcbinputlisting{
  promptbox,
  title={Proposer prompt},
  listing file={prompts/proposer_prompt.txt}
}

\subsection{Validator prompt}

The box below gives the validator prompt. Each validator sees the
full problem and solution, the focused node excerpt, and one proposed error.
The placeholder for the dataset policy is filled with the corresponding block
shown in the following boxes.

\tcbinputlisting{
  promptbox,
  title={Validator prompt},
  listing file={prompts/validator_prompt.txt}
}

\subsection{Dataset-specific policies}

The boxes below contain the complete policy text appended for each benchmark.
H2V uses the rigorous olympiad standard because every claimed proof step is
part of the correctness target; FirstProof-Eval uses the expert-sketch
standard. These dataset-specific policies are applied to every methods including \method, \textsc{Voting} and others.

\tcbinputlisting{
  promptbox,
  title={Putnam100 policy},
  listing file={prompts/policy_putnam100.txt},
  listing options={
    basicstyle=\ttfamily\tiny,
    breaklines=true,
    breakatwhitespace=true,
    columns=fullflexible,
    keepspaces=true,
    showstringspaces=false
  }
}

\tcbinputlisting{
  promptbox,
  title={H2V policy},
  listing file={prompts/policy_h2v.txt},
  listing options={
    basicstyle=\ttfamily\tiny,
    breaklines=true,
    breakatwhitespace=true,
    columns=fullflexible,
    keepspaces=true,
    showstringspaces=false
  }
}

\tcbinputlisting{
  promptbox,
  title={FirstProof-Eval policy},
  listing file={prompts/policy_firstproof_solutions.txt},
  listing options={
    basicstyle=\ttfamily\tiny,
    breaklines=true,
    breakatwhitespace=true,
    columns=fullflexible,
    keepspaces=true,
    showstringspaces=false
  }
}

\end{document}